\documentclass[sn-mathphys,Numbered]{sn-jnl}

\usepackage{graphicx}%
\usepackage{multirow}%
\usepackage{amsmath,amssymb,amsfonts}%
\usepackage{amsthm}%
\usepackage{mathrsfs}%
\usepackage[title]{appendix}%
\usepackage{xcolor}%
\usepackage{textcomp}%
\usepackage{manyfoot}%
\usepackage{booktabs}%
\usepackage{algorithm}%
\usepackage{algorithmicx}%
\usepackage{algpseudocode}%
\usepackage{listings}%
\usepackage{hyperref}%

\theoremstyle{thmstyleone}%
\theoremstyle{thmstyletwo}%

\theoremstyle{thmstylethree}%

\begin{document}

\title[Article Title]{Diagnosis of Skin Cancer Using VGG16 and VGG19 Based Transfer Learning Models}


\author[1]{\fnm{Amir} \sur{Faghihi}}\email{faghihi.a@qut.ac.ir}
\equalcont{These authors contributed equally to this work.}

\author*[1]{\fnm{Mohammadreza} \sur{Fathollahi}}\email{fathollahi@qut.ac.ir}
\equalcont{These authors contributed equally to this work.}

\author*[1]{\fnm{Roozbeh} \sur{Rajabi}}\email{rajabi@qut.ac.ir}
\equalcont{These authors contributed equally to this work.}

\affil*[1]{\orgdiv{Faculty of Electrical and Computer Engineering}, \orgname{Qom University of Technology}, \orgaddress{\street{Khodakaram Blvd.}, \city{Qom}, \postcode{3718146645}, \state{Qom}, \country{Iran}}}


\abstract{Today, skin cancer is considered as one of the most dangerous and common cancers in the world which demands special attention. Skin cancer may be developed in different types; including  melanoma, actinic keratosis, basal cell carcinoma, squamous cell carcinoma, and Merkel cell carcinoma. Among them, melanoma is more unpredictable. Melanoma cancer can be diagnosed at early stages increasing the possibility of disease treatment. Automatic classification of skin lesions is a challenging task due to diverse forms and grades of the disease, demanding the requirement of novel methods implementation. Deep convolution neural networks (CNN) have shown an excellent potential for data and image classification. In this article, we inspect skin lesion classification problem using CNN techniques. Remarkably, we present that prominent classification accuracy of lesion detection can be obtained by proper designing and applying of transfer learning framework on pre-trained neural networks, without any requirement for data enlargement procedures i.e. merging VGG16 and VGG19 architectures pre-trained by a generic dataset with modified AlexNet network, and then, fine-tuned by a subject-specific dataset containing dermatology images. The convolution neural network was trained using 2541 images and, in particular, dropout was used to prevent the network from overfitting. Finally, the validity of the model was checked by applying the K-fold cross validation method. The proposed model increased classification accuracy by 3\% (from 94.2\% to 98.18\%) in comparison with other methods.}

\keywords{Melanoma, Dropout, Overfitting, Convolution neural networks, K-fold, Transfer Learning}



\maketitle

\section{Introduction}\label{sec1}

Cancer includes all types of malignant tumors, often known as neoplasms in medicine. Skin cancer starts from skin cells that make up the main components of the skin. Skin cells grow and divide to form new cells. Then, the cells get old and die, and new ones take their place. A mistake may occur in cell living cycle; new cells appear when they are not needed, and old cells do not die at the appointed time. These extra cells are a mass in form of tissue that is called a tumor. This might happen when one of the body's cells undergoes abnormal growth due to various possible factors, mainly, by continuous exposure to the sunlight, that eventually leads to the development of cancerous tumor, which destructs that part of the body and then spreads to other parts \cite{jain2015computer}. Interest in skin cancer diagnosis and therapy has grown significantly in recent years due to the irreparable damage caused by this kind of cancer and its prevailing widespread occurrence. Skin cancer lesions can be divided into malignant lesions and benign moles.

Among the malignant lesions, melanoma lesions can be considered as one of the most deadly cancerous diseases.  As 70\% of deaths caused by skin cancer  originated from melanoma worldwide. The main damage of skin cancer can be observed as injuries that affect the epidermal layer of the skin to a large extent.

In this regard, timely diagnosis can greatly increase the chances of patient recovery, so several attempts have been dedicated to develop effective methods in order to diagnosis the disease at early stages. Traditional image feature classification techniques have been employed to undertake this crucial task. However, considering that this is the issue of human life at stake, the highest possible accuracy for detection is demanded. For this purpose, deep learning algorithms have been exploited recently to ensure the highest possible accuracy of the results.
In the research performed by Jayalakshmi et al using the PH2 dataset in two classes, they were able to reach an accuracy of 89.3\% by customizing and tuning the CNN model \cite{jayalakshmi2019performance}. In general, the excellency of convolutional neural network in image classification have been approved in various applications, e.g., recognition of car license plates, aerial target tracking. etc, in which, prime performance and accuracy have been obtained \cite{shahidi2022deep, mahdavi2020drone}.

Brindha et al unrevailed  the superiority of CNN algorithm against SVM algorithm in the classification of ISIC image dataset, showing a considerable increase in accuracy from 61\% to 83\%.    \cite{brindha2020comparative}.

Pham and his colleagues were also able to achieve accuracy of 79.5\% and 87\% to classify the ISIC dataset by applying Transfer Learning methods, namely, Reznet50, and InceptionV3, respectively \cite{pham2020improving}. 

Mijwil exploited and compared three different architectures; namely, VGG19, ResNet, and Inception V3 by using ISIC archives between2019 and 2020 with a considerable number of more than 24000 images for skin cancer detection and present the accuracy of each architecture to be 73.11\%, and the best 86.9\%, respectively \cite{mijwil2021skin}.

In addition, Nawaz et al. combined a region-based CNN technique whit support vector Machine (SVM) classifier and employed a dataset based on ISIC-2016 with an increased number of images by using data augmentation step with more than 7000 images for Melanoma classification and obtained accuracy of 89.1\% \cite{nawaz2021melanoma}.

In addition, Alzubaidi and his colleagues were  able to reach 97.5\% accuracy to classify skin lesion images based on deep learning method by implementing a multiphase training scenario and multistage CNN model with the aim to overpass limited number of labeled data for medical application \cite{6}.

In their paper Ashraf et al. considered examination of skin lesion images with the help of deep learning method, including a region of interest segmentation preprocessing and also image augmentation. The result without region of interest segmentation and augmentation was about 81.3\%. An increase to 97.2\% was achieved by applying the segmentation and augmentation \cite{7}. Rafi and coworkers achieved 98.7\% accuracy by applying transfer learning architectures based on Efficient NET-B7, in which extensive image pre-processing such as resizing, conversion, augmentation, and inparticular post scaling step was utilized \cite{rafi2021scaled}.

Lafraxo and coworkers proposed a CNN architecture to recognize Malignant among dermoscopic image, in which, they employed regularization, and particularly, geometric and color augmentations to enlarge the datasets; namely, ISBI (to 18000 image), PH2 (to 2880 images), and MED-NODE (to 1800 images), and were able to achieve the accuracies of 98.44\%, 97.39\%, and 87.77\%, respectively
\cite{lafraxo2022melanet}.

Rasel and his colleagues implemented a transfer learning-based deep CNN model, in which the main ideas had borrowed from the LeNet. Their model included a total of 31 layers and utilized nonlinear variable Leaky ReLU activation function and trained by 250 epochs. They achieved accuracies of 75.5\%, 97.5\%, and 98.33\% for PH2, augmented (rotated) PH2, They achieved accuracies of 75.50\%, 97.50\%, and 98.33\% for PH2, augmented (rotated) PH2, and another dataset i.e. small portion of the images from ISIC archives, respectively \cite{rasel2022convolutional}. 

Hassan et al. comprehensively surveyed the literature to reveal the performance of different optimization algorithms. They also presented 97.3\% (between 92\% up to 98\%) and 99.07\% of accuracies for their deep learning model for ISIC (with ~6000 iterations) and COVIDx (with ~300 iterations), respectively, by using Adam optimizer \cite{hassan2023effect}. 

Also, Hassan et al. achieved a superior accuracy of 97.47\%, employing ResNet50 and Adam optimizer for the classification of retinal optical coherence tomography images with 84495 total number of images \cite{hassan2023enhanced}. 

Alahmadi and coworkers presented a CNN/transformer coupled network, in which both supervised and unsupervised training techniques were utilized. They acquired 95.51\% and 97.11\% accuracy for ISIC and PH2 datasetsimages \cite{alahmadi2022semi}. 

Wu et al. proposed and developed a novel two-stream network, which utilizes a CNN and an extra transformer branch to efficiently capture both local features and global long-range dependencies. They achieved accuracies of 95.78\% (ISIC2018), 93.26\% (ISIC2017), 96.04\% (ISIC2016), and 97.03\% (PH2) for the corresponding datasets. For a better model initialization, they used deit-tiny-distilled-patch16-224 and ResNet34. They also utilized dynamic polynomial learning rate decay \cite{wu2022fat}. 

In this manuscript, our aim is to achieve superior performance and precision through the utilization of a transfer learning model. Our approach involves a novel adaptation and fusion of network architecture and weights, with the primary objective of attaining better detection accuracy while reducing computational burden. Notably, our methodology yields remarkable results in prime detection accuracy without resorting to any data augmentation techniques.

The rest of the paper is organized as follows: Section \ref{methods} discusses methods, model architecture, and used dataset. Then, in section \ref{results} experiments and results are presented along with a discussion on the outputs. Finally, section \ref{conclusion} concludes the paper.

\section{Methods}\label{methods}
In this research, we trained deep neural networks using images of skin lesions. Deep convolutional neural network (CNN) is trained using a dataset of skin lesions images. Adam optimizer and early stopping are used to update the weights of the network. The experiments are performed on Google Colaboratory \cite{hoefler2021sparsity}. The output layer of the proposed model has two different classes, and to prevent the value of loss from increasing, a random removal method has been used.

\subsection{Convolution Neural Network (CNN)}\label{}
Deep learning  methods have broaden the borders of machine learning technology for practical applications. In this class of methods, intermediate layers are employed for data mapping and feature learning, which allows the elimination of non-automatic feature engineering, as the most advantageous distinction of the method. In this regard, for instance, convolution layers operate as the kernel in one of the promising deep learning algorithm, called CNN. Various architectures can be used to process and classify the input image as well as the intermediate feature maps. Later, the pooling layer is used to reduce the size of the feature maps and the network parameters, in which we consider the max-pooling strategy in our model. After the last pooling layer, the fully connected layer is placed, which is mainly responsible for making neural network output into one dimension. As the last layer, the softmax function is placed, which is responsible for binary indexing, 0 and 1, representing the two classes of the images under investigation i.e. normal versus cancerous ones \cite{cocskun2017overview}. The described model is sketched in Figure 1.
\begin{figure}[h]%
	\centering
	\includegraphics[width=0.9\textwidth]{"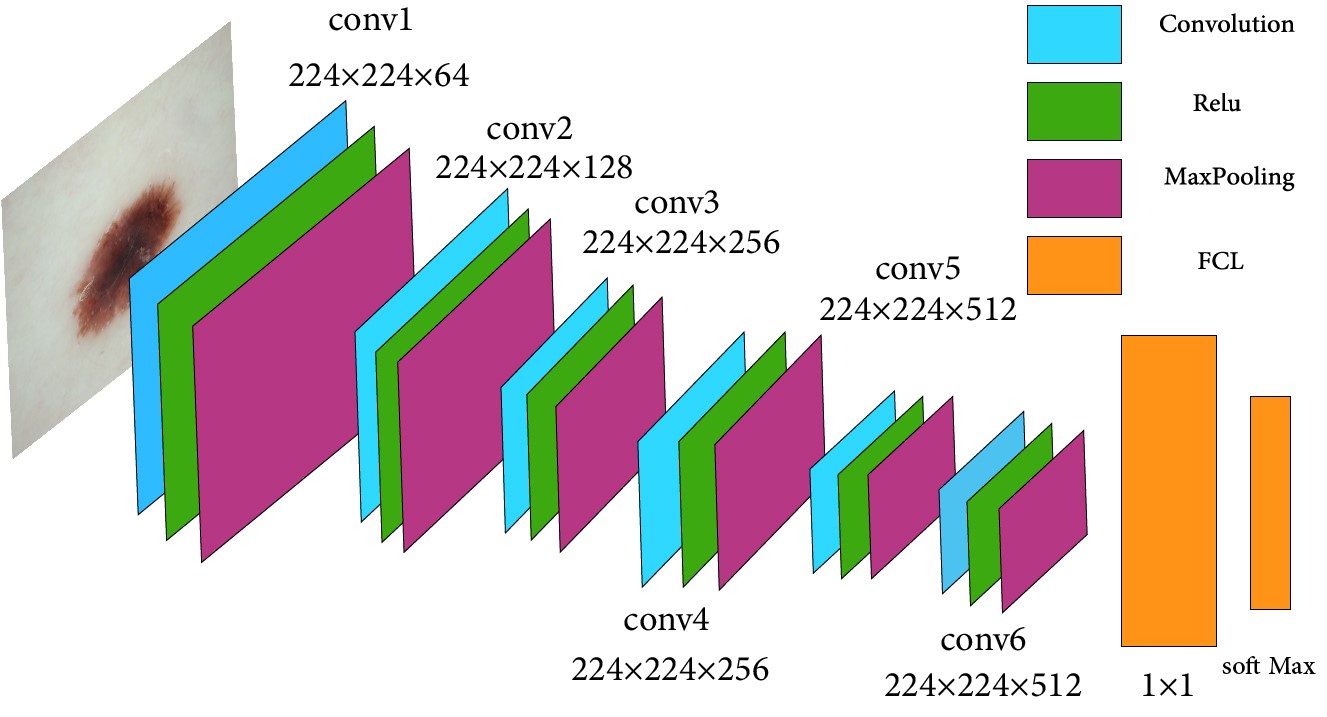"}
	\caption{A view of the customized CNN artichecture.}\label{fig5}
\end{figure}
\subsection{Model architecture}\label{}
The basis of our proposed model lies in the integration of transfer learning principles with the renowned AlexNet architecture, thereby enhancing its performance within the context of our specific dataset. To accomplish this, we embark on a layered approach, supplementing the pre-trained architecture with additional layers through the application of transfer learning techniques. In essence, we amalgamate the weights garnered from the training of the ImageNet dataset using VGG16 and VGG19 architectures with those associated with both the initial three layers and the concluding two layers of our tailored AlexNet variant.

This intricate fusion of weights and architectural components not only imparts a sophisticated depth to our network but also endows it with a broader capacity to discern intricate patterns within the data. Moreover, the amalgamation of these diverse sources of knowledge mitigates overfitting tendencies, a feat that can be attributed to our strategic implementation of the dropout method. This approach introduces a deliberate element of randomness during training, thereby curbing the network's inclination to excessively fit the training data. Through these meticulous steps, our model emerges as a robust solution that not only harnesses the strengths of transfer learning and architectural customization but also effectively manages the delicate balance between model complexity and overfitting prevention \cite{norouzi2019structured}.

The proposed model has been implemented on Google Colab along with the other architectures as reference. We have considered 100 rounds for the training of each network, but at the same time, the early stopping technique has been used to stop the training process to reach the highest available performance in the fastest possible time \cite{hinton2012improving}. Finally, based on our problem of classifying the image set into two classes, the last layer of the neural network has been implemented with 2 neurons. For the core part, ImageNet weights were used for the weights of the proposed network; however, since the images under investigation are medical microscopic images,
we freezed the weights of the first three layers of the network. Usually, to implement transfer learning, it is customary to freeze the last two layers, but due to our examination of  dermoscopy images, to better detect the boundaries of the lesion, we have also kept the weights of the few first layers, and used the ImageNet weights for the rest \cite{gao2018deep,zhang2021squeeze}.

\begin{figure}[tbp]%
	\centering
	\includegraphics[width=0.9\textwidth]{"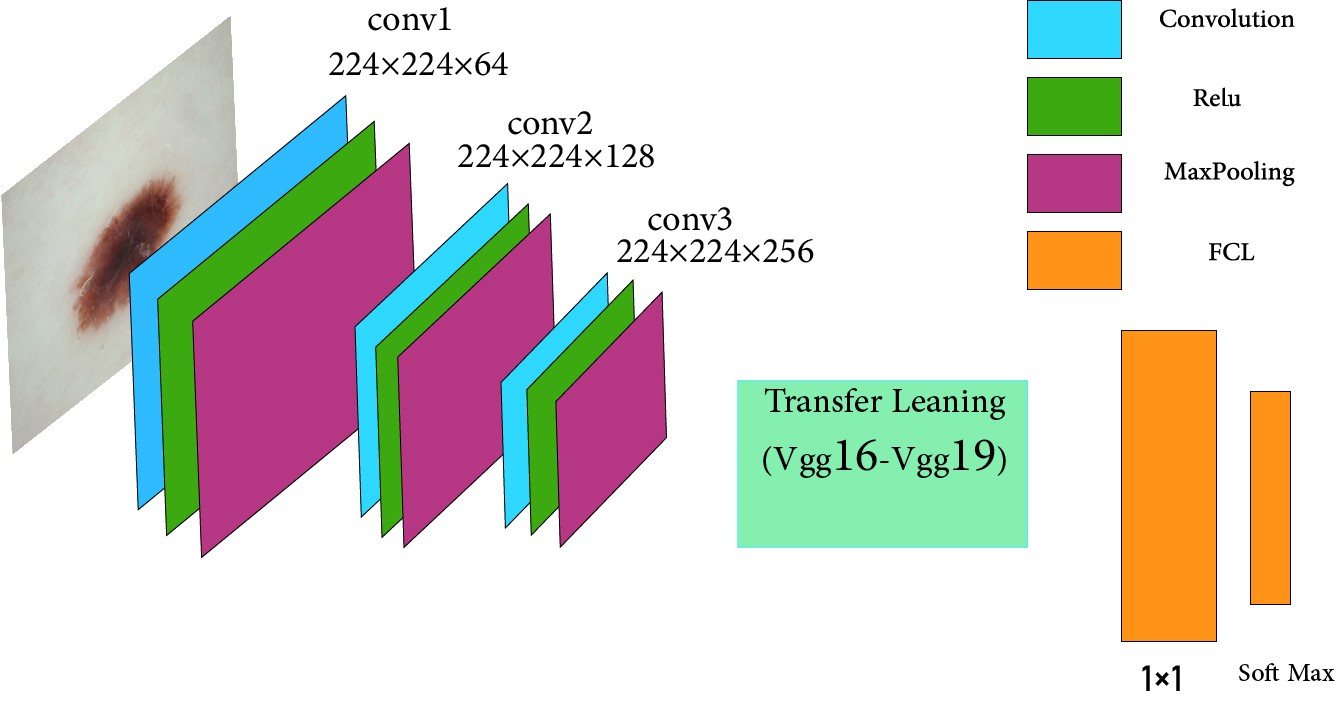"}
	\caption{Proposed transfer learning to Customize CNN}\label{fig2}
\end{figure}

Figure 2 depicts the schematic of the presented CNN model with the application of the customized transfer learning scheme. Particularly, the customization of the network improves the ability of the algorithm to well detect the lesion boundaries, and also increases the speed of convergence and improves the accuracy of model training. 
\subsection{Dataset}\label{}
In the current study, a total of 2541 input images were utilized, comprising 1200 melanoma lesions and 1341 benign mole images. To ensure dataset balance, a reduced number of benign samples were randomly selected \cite{SkinNet23}. For model evaluation, 762 images (30\%) were set aside, while 1779 images were allocated for model training.

The image set is from the International Skin Imaging Collaboration dataset (ISIC) \cite{isic}. This dataset is built from images labeled by Hospital Clinics de Barcelona, Medical University of Vienna, Memorial Sloan Kettering Cancer Center, Melanoma Institute Australia, the University of Queensland, and the University of Athens Medical School. 

In addition, the images of benign and malignant have been taken from the Complete-Mednode-Dataset, published by the Department of Dermatology of the University Medical Center Groningen \cite{mednod} which are combined and used to conduct the present investigation.
 
In general, these lesions are divided into two categories: melanoma lesions and moles Benign (nevus), which are used to detect lesions suspected of malignant melanoma. Figure 3 shows some examples of both cases. In general, the size of the images are $224\times224$ pixels. It is noteworthy that, especially, since each network architecture implementation requires a particular specifications for the input images, we employed a pre-processing function for each case. Some researches conducted in the field tried to generate more images in their data set by cropping or rotating the images or weighting the data. However, in present study, we integrated several datasets to avoid the utilization of duplicated image.

\begin{figure}[tbp]%
	\centering
	\includegraphics[width=0.9\textwidth]{"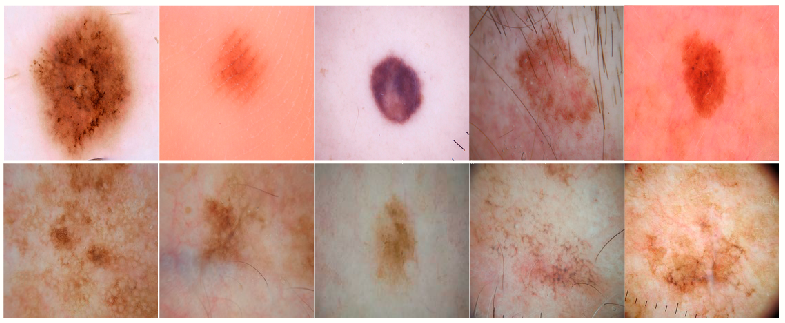"}
	\caption{The first row shows some examples of melanoma lesions, and the second row some examples of harmless moles}\label{fig 3}
\end{figure}

\bigskip
\begin{table}
	\label{tabResults}
	\caption{Comparison results based on evaluation criteria in percent.}
	\centering
	\begin{tabular}{|c|p{3pc}|p{3pc}|p{3pc}|p{3pc}|p{3.5pc}|p{3.5pc}|}                
		\hline
		\verb+Architecture+ & Train Accuracy  & Validation Accuracy & Test Accuracy & k-fold Accuracy & Test Sensitivity   & Test Specificity\\
		\hline
		Modified AlexNet & 97   & 96.5  & 91 & 95.81 & 90.9 & 90.8 \\\hline
		VGG16 & 98   & 97.5  & 92.5 & 97.51 & 96.6 & 95.4 \\\hline
		VGG19 & \textbf{98.7}  & \textbf{98.4}  & \textbf{94.2} & \textbf{98.18} & \textbf{98.08} & \textbf{98.2} \\\hline	
	\end{tabular}
\end{table}
\bigskip

\section{Results and Discussion}\label{results}

In order to present the effectiveness of the proposed model, Figure 4 illustrates the accuracy and performance of our customized transfer learning network based on VGG-16 in comparison with the performance of the reference transfer learning network \cite{sonsare2021cascading}. As can be observed, the detection accuracy increases from 
96.5\% to 97.51\%. Moreover, Figure 5 shows the difference between the use of transfer learning network based on VGG-19 and the model that we have set up. In particular, it is evident that higher accuracy can be reached with the passing of fewer epochs (from 97\% to 98.4\%).

\bigskip
\begin{figure}[tbp]%
	\centering
	\includegraphics[width=0.9\textwidth]{"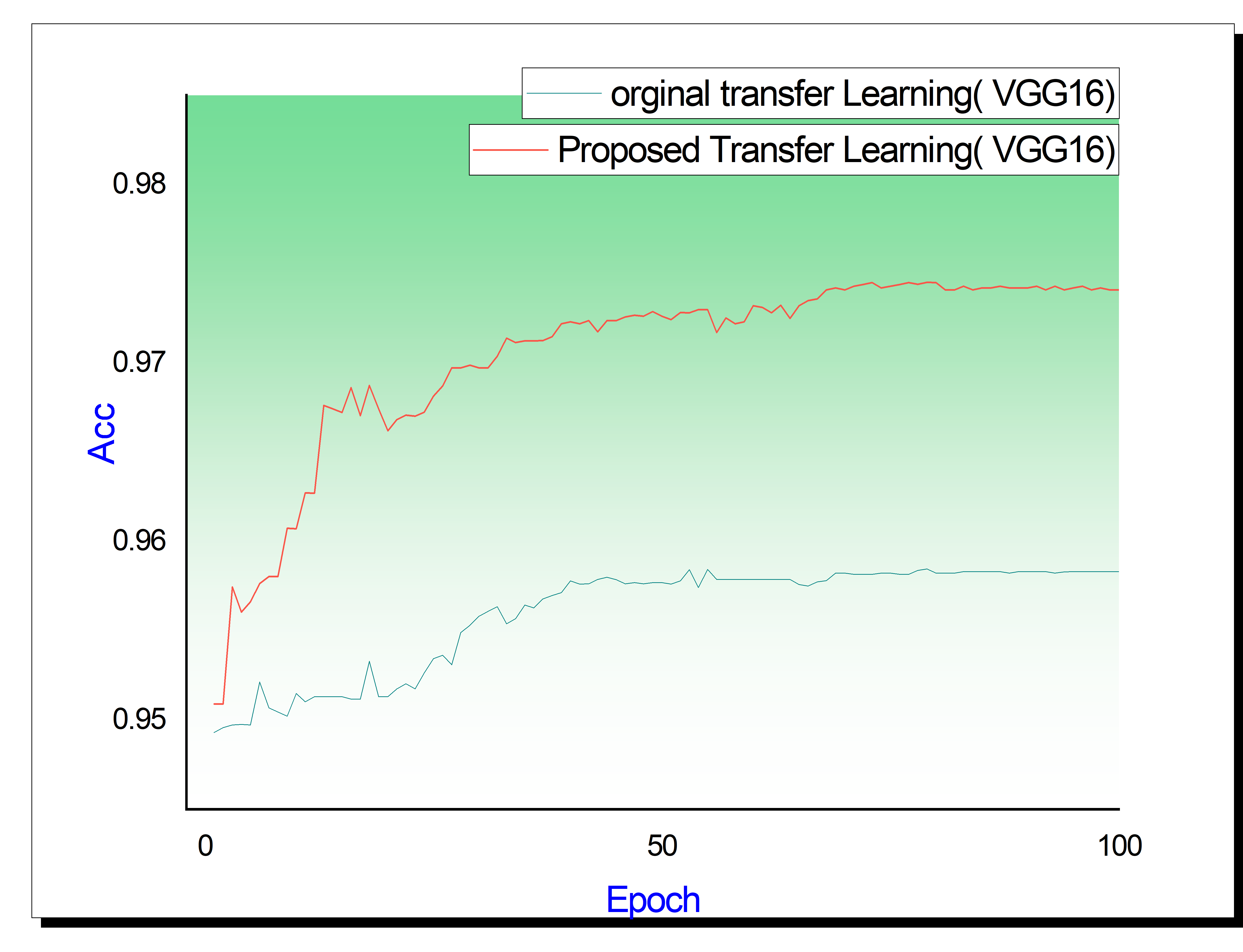"}
	\caption{A comparison chart of Transfer Learning changes based on VGG16, Green: Model performance using normal transfer learning, Red: Model performance using modified transfer learning}\label{fig 4}
\end{figure}

\begin{figure}[h]%
	\centering
	\includegraphics[width=0.9\textwidth]{"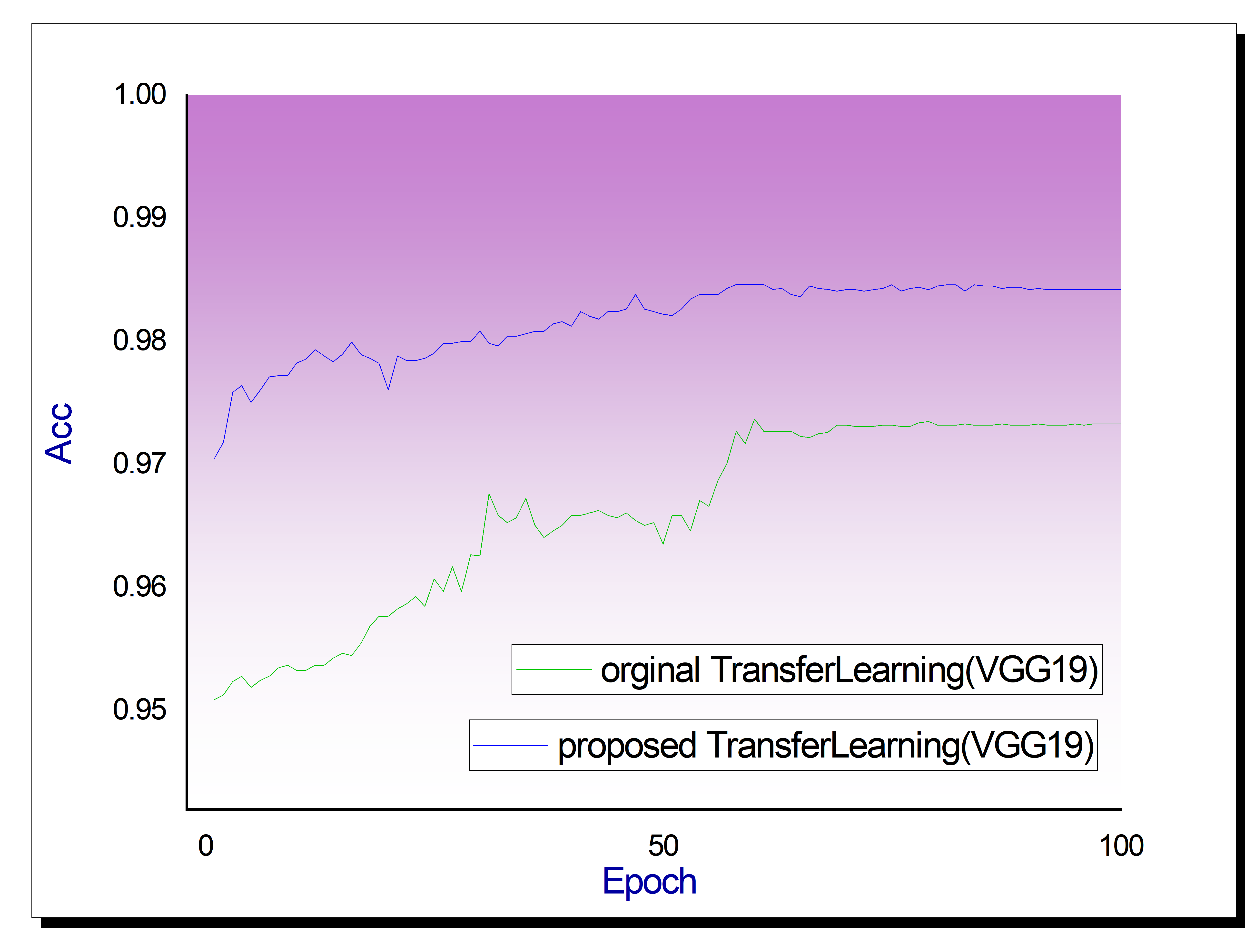"}
	\caption{A comparison chart of Transfer Learning changes based on VGG19, Green: Model performance using normal transfer learning, Blue: Model performance using modified transfer learning}\label{fig 5}
\end{figure}

\subsection{Ablation Study}\label{}
In this experiment, we conducted three separate runs, systematically excluding each of the newly introduced layers, and assessed the resulting impact on the network's performance. The outcomes clearly underscored the remarkable efficacy of the added layers, as the omission of any single layer invariably led to a noticeable decline in accuracy. This compelling evidence highlights the indispensable contribution of each layer to the overall functionality and effectiveness of the network, reaffirming their role in enhancing the model's performance and robustness.

\subsection{Optimizer Selection}\label{}
\begin{figure}[h]%
	\centering
	\includegraphics[width=0.9\textwidth]{"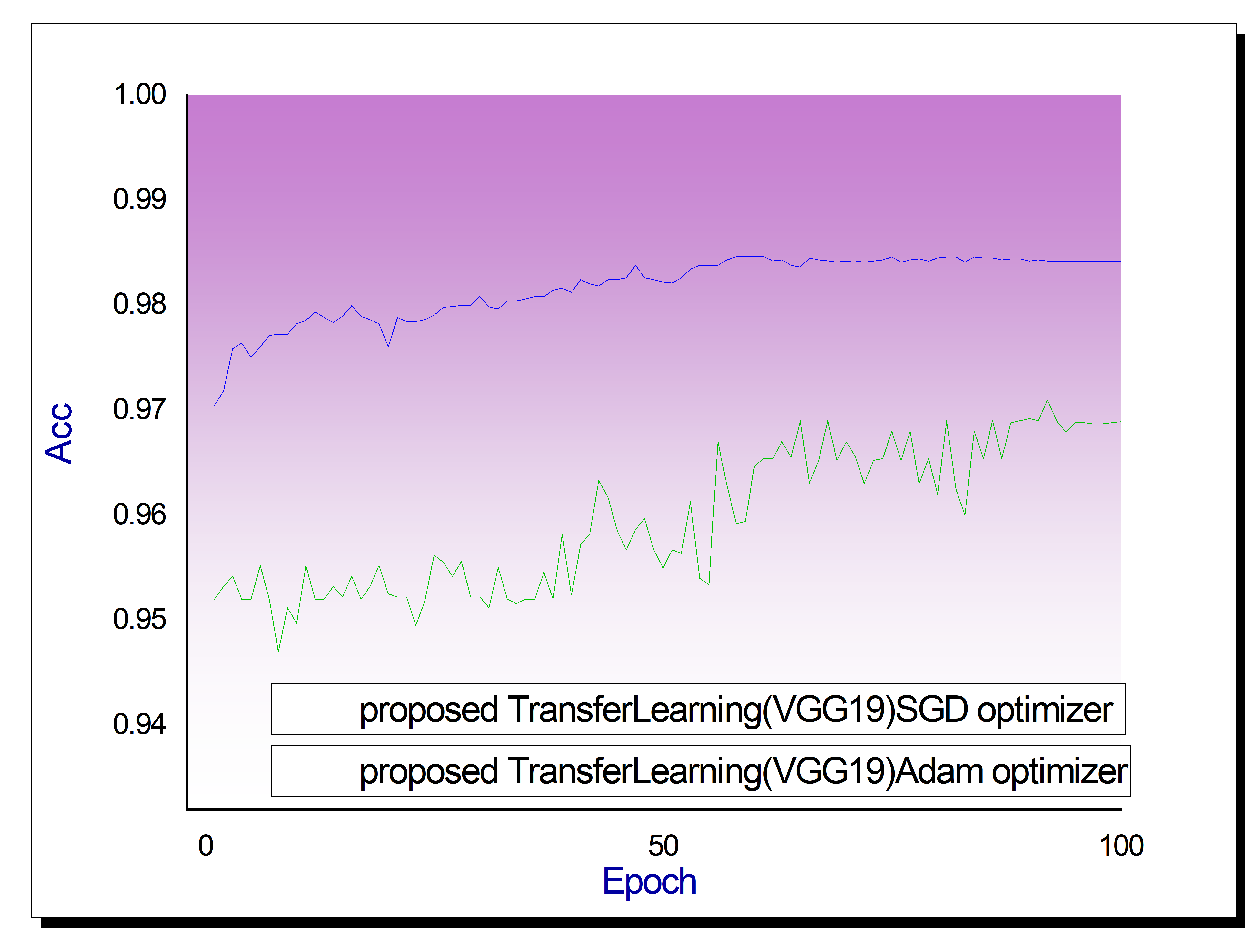"}
	\caption{Comparison of SGD and Adam optimizers.}\label{fig:optimizer}
\end{figure}

In this experiment, we looked at different optimizers. We focused on two specific ones: SGD and Adam. We compared how well they worked and put the results into a graph shown in Figure \ref{fig:optimizer}. From the graph, it's pretty clear that the Adam optimizer performed better than the SGD optimizer. This finding is important because it helps us understand which optimizer is more effective for our specific experiment.

\subsection{K-fold Cross Validation}\label{}
We employed K-fold cross-validation algorithm in order to evaluate and obtain a reliable perdiction 
regarding the true performance of the proposed model to accurately detect the skin lesion between unseen data. Moreover, we can exploit the method to obtain the most optimal values for the hyperparameters of the implemented neural network. We assumed the value of 10 for the K. The average accuracy obtained from the modified VGG-16 and VGG-19 training architecture by employing K-fold method was upper than 97.5\%. Table 1 summarizes the details.

\subsection{Early stopping}\label{}
Two methods have been exploited to prevent overfitting. The first is the dropout setting and the second is the early stopping, as can be seen in the following graphs. By using early stopping, the time required for data processing is significantly reduced.

\begin{figure}[h]%
	\centering
	\includegraphics[width=0.9\textwidth]{"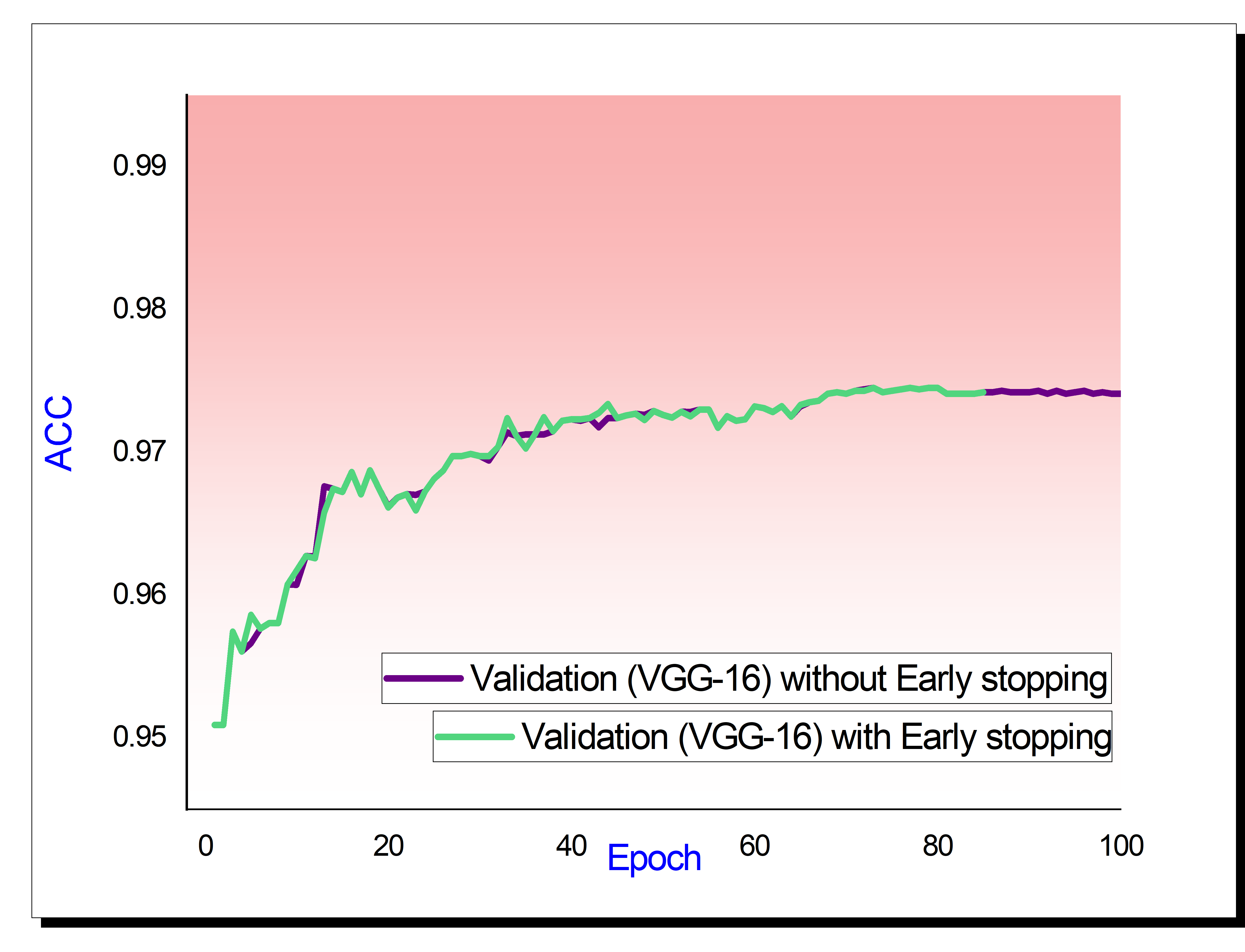"}
	\caption{Comparing the results of the validation data on VGG-16 when the early stop is used with when the early stop is not used.}\label{fig6}
\end{figure}

\begin{figure}[h]%
	\centering
	\includegraphics[width=0.9\textwidth]{"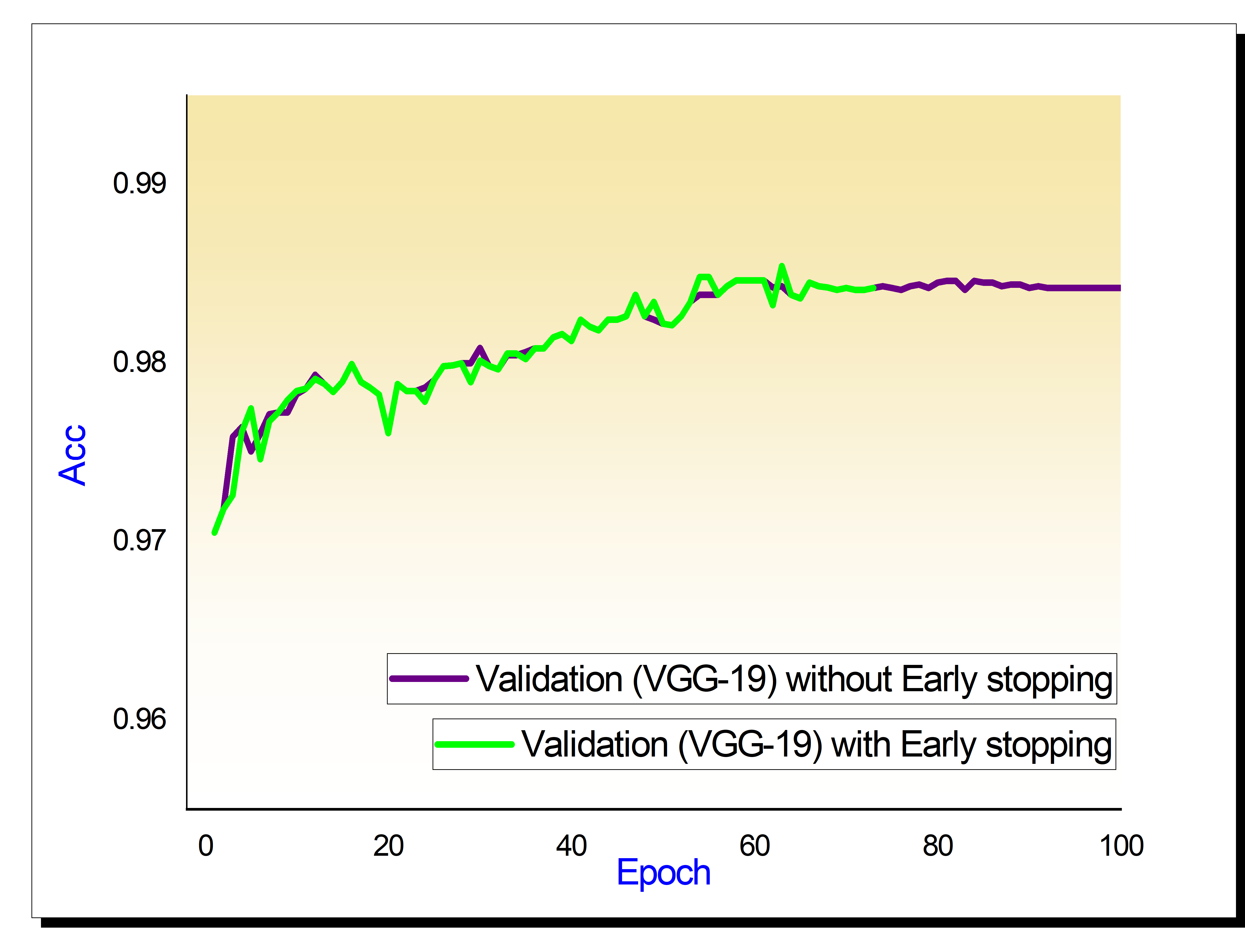"}
	\caption{Comparison of the results of the validation data on VGG-19 when the early stop is used and when the early stop is not used.}\label{fig7}
\end{figure}

Figures 6 and 7 compare the results of the method as advanced by early stopping with the reference cases for the both VGG-16 and VGG-19 based architecture networks. As it can be observed, we avoided extra unnecessary data processing (green plots).

In order to evaluate the performance of the proposed model in comparison with other models, Table 2 summarizes and compares the results of the present study and researches reported in the literature. It can be observed the proposed TL method can achieve significant accuracy with relatively low workflow in comparison with other methods.\\

\begin{table}
	\label{tabResults}
	\caption{Summary of proposed method results in comparison with other methods.}

\begin{tabular}{
		|p{\dimexpr.055\linewidth-2\tabcolsep-1.3333\arrayrulewidth}
		|p{\dimexpr.2\linewidth-2\tabcolsep-1.3333\arrayrulewidth}
		|p{\dimexpr.07\linewidth-2\tabcolsep-1.3333\arrayrulewidth}
		|p{\dimexpr.07\linewidth-2\tabcolsep-1.3333\arrayrulewidth}
		|p{\dimexpr.07\linewidth-2\tabcolsep-1.3333\arrayrulewidth}
		|p{\dimexpr.07\linewidth-2\tabcolsep-1.3333\arrayrulewidth}
		|p{\dimexpr.07\linewidth-2\tabcolsep-1.3333\arrayrulewidth}
		|p{\dimexpr.20\linewidth-2\tabcolsep-1.3333\arrayrulewidth}
		|p{\dimexpr.115\linewidth-2\tabcolsep-1.3333\arrayrulewidth}
		|p{\dimexpr.105\linewidth-2\tabcolsep-1.3333\arrayrulewidth}|
	}
	\hline
	
	No & Architecture Model & \begin{turn}{-90} Train ACC [\%] \space\space\end{turn}	& \begin{turn}{-90} Validation ACC [\%] \space\space\end{turn}  & Test ACC [\%] & Test SE [\%] & Test SP [\%] & Number of Images & Task & Ref. \\ \hline
	
	1 & TL borrowed LeNet + nonlinear variable Leaky ReLU & - & -& 75.5 97.5 98.3 & 88 100 95 & 100 100 100 & 200(PH2) 400(augPH2) ISIC (partially) & Skin lesion & Rasel \cite{rasel2022convolutional} \\ \hline

	2 &	ResNet-based CNN + Adam optimizer & - 98.8 & - & 97.3 99.1 & 97.7 99.5 & 96.9 
	98.6 & 2594(ISIC) 13413(COVIDx) & Skin cancer CT image COVID &	Hassan \cite{hassan2023effect} \\ \hline
	
	3 & Modified ResNet + SGD/Adam optimizer & - & -	& 97.5 & 98.4 &	96.2 &	84495(OCT2017)& Retinal tomography image & Hassan \cite{hassan2023enhanced} \\ \hline
	
    4 & CNN coupled transformer + supervised unsupervised training & - & - & 95.9 95.9 97.1 &  88.2 94.8 93.4 & 97.1 97.7 97.8 & 3694(ISIC2018) 2750(ISIC2017) 200(PH2) & Skin lesion & \begin{turn}{-90} Alahmadi \cite{alahmadi2022semi} \space\space\end{turn} \\ \hline

    5 & transformer CNN model feature adaptive deit-distilled-path initialization & - & - & 95.8 93.3 96.0 97.0 & 91.0 83.9 92.6 94.4 &	97.0 97.3 96.0 97.4	& 2594(ISIC2018) 2750(ISIC2017) 1279(ISIC2016) 200(PH2) & \begin{turn}{-90} Melanoma \space\space\end{turn} & Wu \cite{wu2022fat} \\
    \hline
    
    6 & Proposed Method & 98.7 & 98.4 & 94.2 & 98.1 &	98.2 & 2541 (ISIC2020 and MED-NODE (partially)) & \begin{turn}{-90} Melanoma \space\space\end{turn} & This paper \\
    \hline
\end{tabular}
\end{table}

As the wrap-up, the examination of skin lesion images is a challenging task due to high degree of similarity between the images; however, based on the modification that was introduced into the transfer learning method, we could beneficially increase the accuracy of the detection. Table 2 summarizes the superiority of the proposed model in present paper over the reference studies. The table present the average values.

\section{Conclusion}\label{conclusion}
In recent years, the adoption of the transfer learning method has gained considerable attention among researchers, owing to its advantages in enhancing model performance. However, it remains imperative to tailor the network's training to suit the specifics of each dataset. This paper has delved into this intricate landscape, striving to enhance the capabilities of deep networks by meticulously adjusting the layer configuration and weight distribution to align with the demands of detecting lesion-affected regions within images. As a testament to our endeavors, achievements have been realized, with accuracy levels reaching 92.5\% for the VGG-16 architecture and an even more impressive 94.2\% for the VGG-19 architecture. We also used k-fold cross-validation methodology, which ensures a robust and unbiased assessment of our proposed model's performance. Employing k-fold the accuracy of 97.51\% for the VGG-16 architecture and 98.1\% for the VGG-19 architecture have been achieved.

Looking ahead, our work opens paths for future exploration. It would be worthwhile to consider the impact of different pre-trained architectures, as well as to explore how varying degrees of fine-tuning could further enhance the model's efficacy. Additionally, while our study showcases promising outcomes, it's essential to acknowledge its limitations. As with any methodology, there are constraints to consider, such as the potential for overfitting in more complex datasets or the challenges associated with domain shifts. Addressing these shortcomings and expanding upon the strengths of our approach will undoubtedly pave the way for the continued evolution of accurate and efficient lesion detection methods.

\section*{Declarations}

\begin{itemize}
	\item Conflict of interest: The authors declare that they have no conflict of interest.
	\item Availability of data and materials: The datasets analysed during the current study are available in the ISIC 2020 repository \href{https://challenge2020.isic-archive.com/}{https://challenge2020.isic-archive.com/}, and MED-NODE repository \href{https://www.cs.rug.nl/~imaging/databases/melanoma_naevi/}{https://www.cs.rug.nl/~imaging/databases/melanoma\_naevi/}
\end{itemize}

\bibliography{ref.bib}

\end{document}